\documentclass[runningheads]{llncs}
\usepackage[T1]{fontenc}
\usepackage[misc]{ifsym}
\usepackage{amsmath} 
\usepackage{amsfonts}
\usepackage{graphicx}
\usepackage{hyphenat}
\usepackage[implicit=true]{hyperref}   
\usepackage{accessibility}   
\providecommand{\Description}[1]{}

\newcommand{\myparagraph}[1]{\textbf{#1}}
\newcommand{\etal}{et~al.}

\begin{document}

\title{Automated Detection of Abnormalities in Zebrafish Development}

\author{%
  Sarath Sivaprasad\inst{1} \Letter%
  \and
  Hui\hyp{}Po Wang\inst{1}%
  \and
  Anna\hyp{}Lisa Jäckel\inst{2}%
  \and
  Jonas Baumann\inst{2}%
  \and
  Carole Baumann\inst{2}%
  \and
  Jennifer Herrmann\inst{2}%
  \and
  Mario Fritz\inst{1}%
}


\authorrunning{S. Sivaprasad et al.}

\institute{CISPA Helmholtz Center for Information Security, Saarbrücken, Germany\\
\email{\{sarath.sivaprasad, hui.wang, fritz\}@cispa.de}
\and
Helmholtz Institute for Pharmaceutical Research Saarland, Saarbrücken, Germany\\
\email{\{anna-lisa.jaeckel, jonas.baumann, carole.baumann, jennifer.herrmann\}@helmholtz-hips.de}}

\maketitle             
\begin{abstract}
Zebrafish embryos are a valuable model for drug discovery due to their optical transparency and genetic similarity to humans. However, current evaluations rely on manual inspection, which is costly and labor-intensive. While machine learning offers automation potential, progress is limited by the lack of comprehensive datasets.
To address this, we introduce a large-scale dataset of high-resolution microscopic image sequences capturing zebrafish embryonic development under both control conditions and exposure to compounds (3,4-dichloroaniline). This dataset, with expert annotations at fine-grained temporal levels, supports two benchmarking tasks: (1) fertility classification, assessing zebrafish egg viability (130,368 images), and (2) toxicity assessment, detecting malformations induced by toxic exposure over time (55,296 images).
Alongside the dataset, we present the first transformer-based baseline model that integrates spatiotemporal features to predict developmental abnormalities at early stages. Experimental results present the model’s effectiveness, achieving 98\% accuracy in fertility classification and 92\% in toxicity assessment. These findings underscore the potential of automated approaches to enhance zebrafish-based toxicity analysis. Dataset and code will be available at: \href{https://github.com/sarathsp1729/Zebrafish-development}{https://github.com}.

\end{abstract}
\section{Introduction}
Zebrafish embryos have emerged as a powerful model organism in drug discovery due to its genomic similarity to humans~\cite{zon2005vivo,lieschke2007animal}.
Their optical transparency further enables high-throughput imaging, allowing experts to conduct drug discovery and investigate the effect of novel compounds at scale, which paves a new way for more reliable and cost-effective analyzes~\cite{macrae2015zebrafish}. 
However, evaluating pharmacological interventions on the development of zebrafish embryos remains a labor intensive task. Unlike classical classification tasks, the challenge arises from accurately detecting subtle developmental abnormalities that manifest over time in image sequences~\cite{rennekamp201515}, which often relies on human expertise~\cite{rihel2010zebrafish}.

\begin{figure}[tb]
\centering
\includegraphics[width=0.8\textwidth]{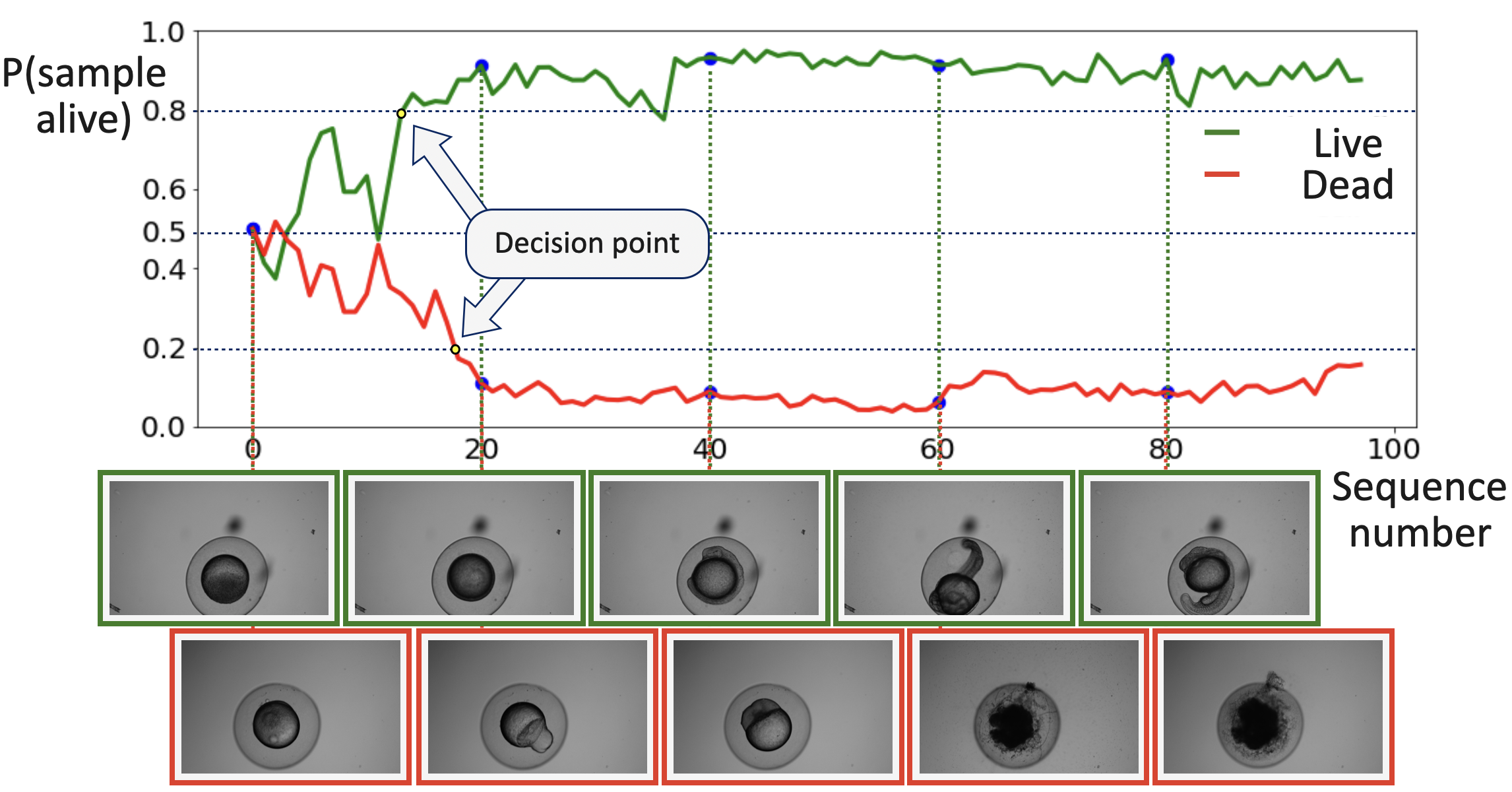}
\Description{Two time-series curves for zebrafish development: green line rises sharply, indicating normal growth; red line remains low, marking anomalous development. Shows model discrimination over time.}

\caption{Illustration of model predictions for two developmental sequences. The red line denotes predicted anomalous development, while the green line represents predicted normal development.} 
\label{teaser}
\end{figure}

Given this challenge, significant early efforts have focused on automating toxicity detection using zebrafish embryos. Traditional approaches often rely on static images, failing to capture the temporal resolution necessary for analyzing dynamic developmental processes~\cite{mikut2013automated}. More recently, the machine learning (ML) community has explored various methods to incorporate spatiotemporal information in fields such as anomaly detection~\cite{iakovidis2018detecting,zhao2021good} and video classification~\cite{karpathy2014large}. However, unlike these applications, clinical studies involving zebrafish embryos lack comprehensive datasets to support such advancements~\cite{rennekamp201515}. This scarcity has hindered the development of an end-to-end solution for accurate toxicity detection, highlighting a critical gap in the field.

To address this, we introduce a comprehensive large-scale dataset of high-definition images capturing the temporal progression of zebrafish embryonic development. The dataset is carefully collected under control conditions and exposure to compound 3,4-dichloroaniline, resulting in around $20$K high-quality images with per-frame annotations. The dataset enables two key tasks: (1) fertility detection, which assesses zebrafish egg viability, and (2) toxicity assessment, which identifies developmental abnormalities by toxic exposure.

Beyond the dataset, we also introduce the first transformer-based baseline model designed to integrate spatiotemporal relationships for the task. It explores the modern transformer architectures~\cite{vaswani2017attention} to model the correlation between different image patches across both spatial and temporal dimensions. The proposed model achieves impressive accuracy with 98\% accuracy on fertility detection and 92\% accuracy on toxicity assessment, highlighting the great potential toward automatic drug discovery. We summarize our three contributions as follows.

\begin{itemize}
    \item We propose a dataset of 130,368 images for fertility detection and 55,296 images for toxicity assessment, with expert temporal annotations, of zebrafish embryonic development under pharmacological interventions.
    \item We introduce a transformer-based model that effectively processes spa\-tio\-tem\-po\-ral features in image sequences, achieving high accuracy in detection (98\% on fertility detection and 92\% on toxicity assessment). The fine grained temporal annotations enable rigorous benchmarking of models for early anomaly detection, for improving intervention studies.
    
\end{itemize} 

\section{Related Work}
\label{sec:related_work}

\myparagraph{Utilizing zebrafish as a model organism:} zebrafish have become an indispensable model in biomedical research due to their rapid embryonic development, optical transparency, and remarkable genomic similarity to humans~\cite{zon2005vivo,lieschke2007animal}. 
In addition to these foundational studies, recent reviews have highlighted how advances in high-throughput imaging and automated data analysis further enhance the zebrafish’s role in toxicity testing~\cite{horzmann2018making}.
Approaches that integrate time-resolved gene expression data into toxicity models have underscored the significance of using zebrafish by revealing insights into the molecular underpinnings of toxic responses in their embryos~\cite{schunck2024integrating}. These advancements further justify the continued inclusion of zebrafish in large-scale screening initiatives, like the United States Environmental Protection Agency’s Toxicology Testing Phase I~\cite{padilla2012zebrafish} and Phase II screens~\cite{truong2014multidimensional}.

\begin{sloppypar}

\myparagraph{High-throughput imaging and annotation:} High-definition imaging platforms have been used to capture zebrafish development at high resolution~\cite{macrae2015zebrafish}. 
A widely used image dataset for analyzing toxicity in zebrafish is ToxCast~\cite{padilla2012zebrafish}. The dataset includes $309$ unique zebrafish embryos, each imaged once at $6$ days post-fertilization and annotated for viability, hatching status, and malformations. However, this dataset lacks both the temporal span and the scale required to train large-scale models. A larger dataset, comprising 203,520 zebrafish embryos, is presented in~\cite{zhang2017new}. This dataset contains video recordings captured for $15$ minutes, annotated for behavioral metrics. Nevertheless, the dataset lacks the temporal annotation for automating the early detection of developmental anomalies. Our dataset gives up to 8 hours (fertility) / 48 hours (toxicity) monitoring data and expert frame annotation enabling early detection benhmarking. Mikut \etal \cite{mikut2013automated} provides a comprehensive survey of the proposed datasets for analyzing of zebrafish development including those annotated for cell tracking and heartbeat detection. There remains a need for a large-scale dataset focused on early development tracking, with temporal annotation of anomalous behavior.
\end{sloppypar}

\myparagraph{Automating developmental toxicity detection:} Various methods have been developed to automate toxicity detection using these datasets.
Earlier methods employed targeted parameters such as velocity, distance traveled, and rule-based scripts~\cite{kokel2010rapid,cachat2011three}.
Key tasks for analyzing drug toxicity have been image-based heartbeat detection~\cite{5872696,spomer2012high} and movement detection from videos~\cite{cario2011automated}.
While these tasks provide valuable information, they do not replace comprehensive temporal analyses of full-image data, for accurate detection of developmental anomalies.

Earlier automation methods trained machine learning models, such as randomized trees, on pixel-based image descriptors~\cite{alshut2010methods}. 
Later approaches leveraged Convolutional Neural Networks for robust developmental anomaly detection~\cite{rennekamp201515,dong2015deep,xu2017zebrafish,tyagi2018fine}. Although transformer models~\cite{dosovitskiy2020image} have been applied to zebrafish data~\cite{javanmardi2023unsupervised,xu2024technical} a pipeline for the detection of toxicity in the early stages of development has not yet been proposed. We bridge this gap by proposing a large-scale dataset with fine-grained annotations and by training an end-to-end transformer-based model for early toxicity detection.

\section {Dataset}
The dataset presented here was collected to develop machine learning models that enable drug discovery using zebrafish as a model organism. Hence, we formulated two key tasks: (1) fertility detection, to assess egg viability, and (2) toxicity assessment, to identify developmental malformations. Eggs were produced through the natural spawning of zebrafish and individually allocated into wells of a standard $96$-well plate. This experimental setup allows the study of the effects of various test substances at different concentrations, with appropriate control conditions. We use the reference compound 3,4-dichloroaniline to study its effects on development and to train the machine learning model. Employing this well-characterized compound establishes a standardized baseline, ensuring that our dataset remains broadly applicable.

\myparagraph{Fertility detection Task:} Data for the task were collected over 14 experimental runs with no added substance. In each run, every well (i.e., each egg) was imaged sequentially over an $8$-hour period. Images were acquired at $5$-minute intervals, resulting in $97$ images per sequence. In total, the fertility classification dataset comprises $\left(\frac{8 \times 60}{5} + 1\right) \times 14 \times 96 = 130,368 \text{ images}$, that is $1344$ sequences.

Each image in the sequence is annotated with one of three labels: `alive', `unsure', or `unfertilized'. Additionally, a flipping point is annotated to mark the transition from an `unsure' state to one of the definitive outcomes. The label of the $97^{\text{th}}$ image is the sequence label. The final dataset exhibits a $5:4$ ratio for the two definitive sequence labels (`alive' and `unfertilized’).

\myparagraph{Toxicity Assessment Task:}
For toxicity assessment, the focus is on detecting the adverse effects of the toxic reference compound, 3,4-dichloroaniline, on zebrafish embryonic development. Each image is annotated, enabling benchmarking of models for the early detection of developmental abnormalities. Since the task requires long-term monitoring, each well on the 96-well plate is observed over a $48$-hour period, with images captured at $15$-minute intervals. Consequently, each sequence contains 192 images ($4$ images per hour over $48$ hours). With data collected from three such runs, the toxicity assessment dataset comprises $192 \times 96 \times 3 = 55\,296$ images, corresponding to $288$ sequences. Each image is analyzed and annotated with one of the following labels: `alive', `sublethal effect', `lethal effect', or `not fertilized', allowing for fine-grained analysis of toxic effects over time. Additionally, the sequence is labeled as `alive' or `anomalous' based on the overall outcome.

\myparagraph{Data collection, accessibility and ethical considerations:} All imaging was performed using a high-definition microscope that captures images at a resolution of $1344\times820$ pixels. The experimental setup ensured that each well maintained consistent lighting and focus conditions. Expert annotators provided the labels based on standardized guidelines, with quality control checks performed to ensure annotation consistency and data integrity. The complete dataset, along with detailed documentation on devices used and annotation guidelines, will be made publicly available upon publication of this manuscript. As zebrafish embryos are widely used as a model organism and do not require invasive procedures, ethics approval was not necessary for this study. 

\begin{figure}[t]
\centering
\includegraphics[width=0.85\textwidth]{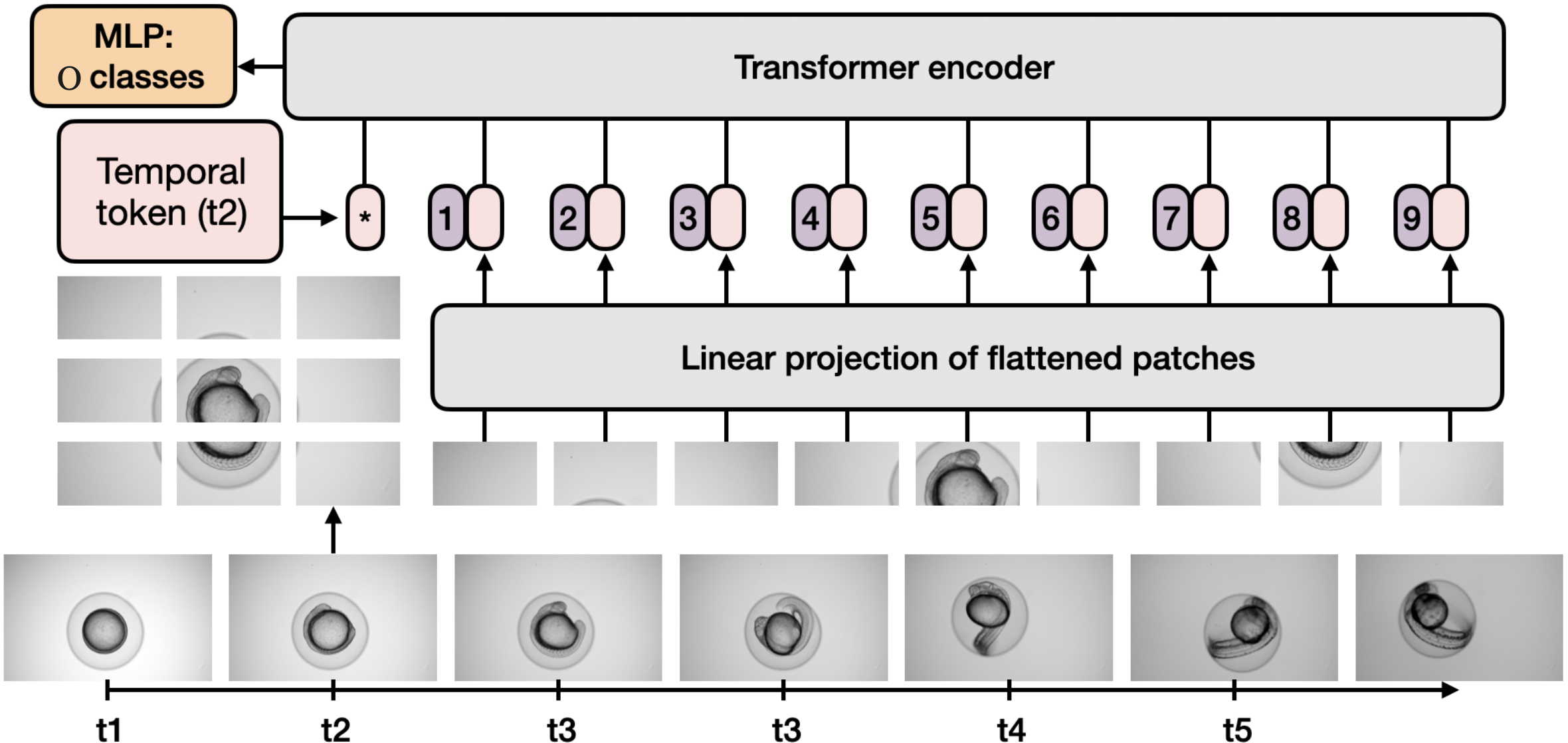}
\caption{Overview of the model architecture. Input images are divided into non-overlapping patches, encoded with patch and temporal embeddings, then processed through a transformer encoder and an MLP classification head.} 
\Description{Flow diagram of the proposed transformer: image frames split into patches, encoded and temporal embeddings added. These tokens pass transformer encoder, MLP head outputs class score.}

\label{model}
\end{figure}

\section{Model}
\label{sec:baseline}

The model is designed to process sequences of microscopic images and capture the dynamic development of zebrafish embryos. Each sequence consists of $N$ images, each representing a snapshot of a zebrafish embryo captured at uniform time intervals. These sequences are used to train a transformer-based model that leverages spatiotemporal features to predict developmental abnormalities.

Let $\mathbf{X} = \{\mathbf{x}_1, \mathbf{x}_2, \dots, \mathbf{x}_N\}$ represent an image sequence, where $\mathbf{x}_t \in \mathbb{R}^{H \times W \times C}$ is the $t$-th image in the sequence, with $H$, $W$, and $C$ denoting the height, width, and number of channels of the image, respectively. Although the original images have a resolution of $1344\times820$ we resize them to $H = 224, W = 224$ ($C=3$).

Each image $\mathbf{x}_t$ captures the developmental state of a zebrafish larva at time $t$. We employ a Vision Transformer (ViT)~\cite{dosovitskiy2020image} architecture adapted for spatiotemporal processing. The model processes each image $\mathbf{x_t}$ along with an encoding for the time instance $t$. For each image $\mathbf{x}_t$, the embedding layer combines three components (Figure~\ref{model}):
\begin{itemize}
    \item Patch Embeddings: Each image $\mathbf{x}_t$ is divided into non-overlapping patches of size $P \times P$. These patches are linearly projected into a $d$-dimensional space, where $d$ is the hidden dimension of the transformer. Let $\mathbf{P}_t \in \mathbb{R}^{M \times d}$ represent the patch embeddings for the $t$-th image, where $M = \frac{H \times W}{P^2}$ is the number of patches per image. We choose $P=16$ and $d=768$ in our experiments, resulting in $M=196$ patches, each embedded as a $768$-dimensional vector.
    \item To encode the spatial location of each patch within the image, a learnable spatial positional embedding $\mathbf{E}_{\text{spatial}} \in \mathbb{R}^{M \times d}$ is added to the patch embeddings. This ensures that the model can distinguish between patches based on their spatial positions.
    \item A class token ($\mathbf{c}$)—a learnable embedding that aggregates global information for classification—is prepended to the patch tokens. To encode the temporal position of each image in the sequence, a learnable temporal positional embedding is added to the class token. This vector is shared across all patches of $\mathbf{x}_t$: $\mathbf{E}_{\text{temporal}} \in \mathbb{R}^{d}$. Post training the model has $N$ unique tokens one for each time step. That is, for a given, $\mathbf{x}_t$ the temporal token is assigned a unique embedding based on the value of $t$.
\end{itemize}

The final embedding $\mathbf{z_t}$ for the $t$-th image is computed as:

\begin{equation}
\mathbf{z}_t = \operatorname{concat}\!\Big(\mathbf{c} + \mathbf{E}_{\text{temporal}}[t],\, \operatorname{concat}_{i=1}^{M}\big(\mathbf{P}_t^{(i)} + \mathbf{E}_{\text{spatial}}^{(i)}\big)\Big)
\end{equation}

where $\mathbf{P}_t^{(i)}$ is the embedding of the $i$-th patch of image $\mathbf{x}_t$, $\mathbf{E}_{\text{spatial}}^{(i)}$ is the corresponding spatial positional embedding, $\mathbf{c}$ is the class token, and $\mathbf{E}_{\text{temporal}}[t]$ is the temporal embedding for time step $t$. $\mathbf{z}_t \in \mathbb{R}^{(M+1) \times d}$ is the representation of image $\mathbf{x}_t$, enriched with spatio-temporal information.

\myparagraph{Transformer encoder and classification head:}
The transformer encoder consists of a stack of identical blocks. Each block first applies layer normalization to the input tokens and then computes multi-head self-attention, which captures global contextual relationships among the tokens. The resulting output is added to the original input via a residual connection. Next, the block processes the tokens with a feed-forward network (again with layer normalization and residual connections), further refining their representations. This sequence of operations enables the encoder to iteratively build a rich, global representation of the spatio-temporally enriched input tokens. After processing through $12$ such blocks, the encoder outputs $\mathbf{h}$-a tensor of the shape $\mathbb{R}^{(M+1) \times 768}$.  

The classification head is a linear layer that processes the representation vector $\mathbf{h}$ to predict the developmental outcome. The output is passed through an activation (A: Sigmoid for fertility, softmax for toxicity), yielding $\mathbf{y} = \text{A}(\mathbf{W} \mathbf{h} + \mathbf{b})$, where $\mathbf{W} \in \mathbb{R}^{o \times d}$ and $\mathbf{b} \in \mathbb{R}^o$ are learnable parameters, and $\mathbf{y} \in \mathbb{R}^o$ represents the predicted probabilities for the $o$ classes. For the task of fertility detection, we set $o=1$ with Sigmoid activation, where the value of the prediction being $0$ indicates `unfertilized' and $1$ indicates `fertilized'. Thus, the prediction for an image $\mathbf{x}_t$ (denoted by $\mathbf{y_t}$) is the probability of the sample being fertile at time $t$ (Figure \ref{teaser}). For task (2), toxicity assessment, we set $o = 2$: corresponding to the two classes `alive' and `anomalous'. We employ softmax activation with $o=2$ to enable future integration of additional classes with fine grained sublethal effects. 

This baseline model enables image-level prediction while incorporating temporal information. During inference, the model processes one image at a time and makes predictions based on both the image content and its temporal index.

\begin{figure}[t]
\centering
\includegraphics[width=0.9\textwidth]{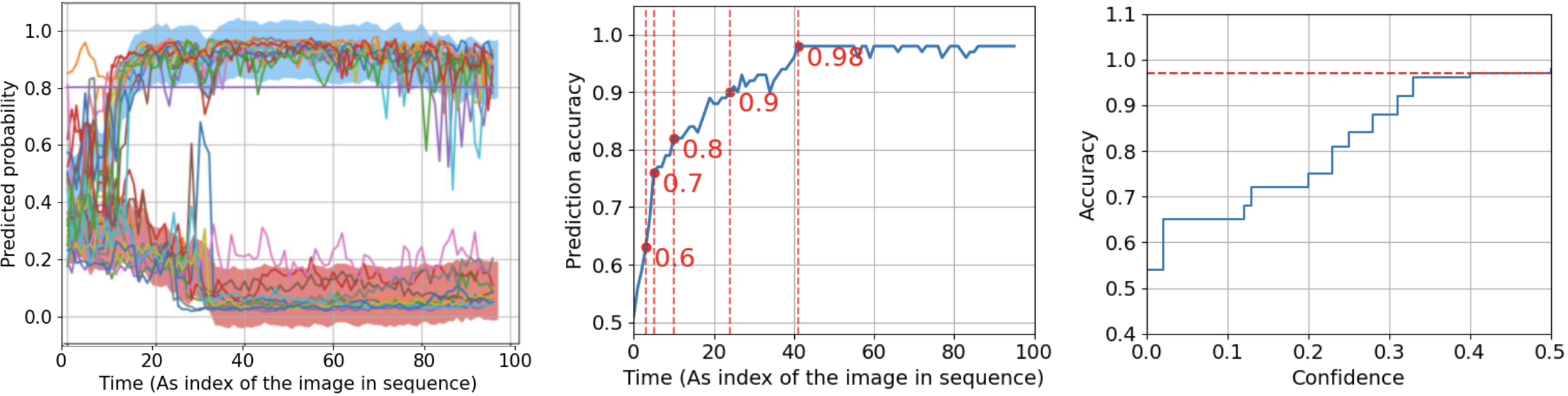}
\caption{The figure shows, from left to right, first examples of model output for random $20$ test sequences. The second figure shows the prediction accuracy changing with time and the third plot shows the confidence calibration of the model.} 
\Description{Three-panel composite: (a) 20 example probability-over-time traces for test sequences; (b) accuracy curve climbing to 98\% as model see later frames; (c) calibration diagram comparing confidence to observed accuracy.}

\label{result1}
\end{figure}

\section{Experiments and Results}
\label{sec:exp}

In this section, we present the experimental evaluation of our transformer-based model on two critical tasks: fertility detection and toxicity assessment. We describe the experimental setup, outline the evaluation, and report the results.

\myparagraph{Training strategy:} We adopt the hyperparameters from a previous state-of-the-art anomaly detection model~\cite{mirzaei2022fake}. For both datasets and tasks, we train the model using the Adam optimizer with a learning rate of $4 \times 10^{-4}$, weight decay of $5 \times 10^{-5}$, and a dropout rate of $0.2$. The training process involves minimizing (1) binary (for fertility detection) and (2) categorical (for toxicity assessment) cross-entropy loss over the labeled dataset.

\myparagraph{Results and discussion:} For the fertility detection task, our dataset comprises $130,368$ images, which form $1344$ sequences. We split the dataset into training, validation, and test sets in a $70:15:15$ ratio, yielding 942 training sequences and 201 sequences each for validation and testing. The model predicts a probability $y_t$ for each image, indicating the likelihood that the sample is fertile. The validation set is used to select the best model during training and to determine the optimal kernel size for smoothing the sequence of predictions. We found that a sliding window size of $13$ is optimal for a moving average filter.


\begin{sloppypar}
For evaluating image-level prediction (i.e., the accuracy of \(y_t\)) we consider only test images with definitive labels—either ‘fertile’ or ‘unfertilized’—and exclude those labeled ‘unsure’. The resulting binary classification accuracy is 89.23 \%. Future models can improve on this and leverage the fine grained classes. However, to evaluate the capability of the model to automate the detection of viable samples, we need to evaluate the accuracy of predicting the state of an entire zebrafish embryo (i,e,. the whole sequence). For this we evaluate (a) the accuracy at a specific time $t$, and (b) the accuracy when the confidence ($c_f$) first reaches a predefined threshold value. The confidence $c_f$ is measured as $2 \times |y_t - 0.5|$. Note that $y_t$ can be understood as the probability of the sample being alive at time $t$. Figure~\ref{result1} illustrates sample outputs along with both analyses. 
\end{sloppypar}

Our analysis indicates that achieving a maximum accuracy of 98\% requires waiting for $41$ time steps in the sequence, which corresponds to $41 \times 5$ minutes: \~{}$3.4$ hours. Although this 3.4-hour duration falls within the typical $8$-hour lab session, it is substantially longer than the $1.3$ hours in which human experts can accurately predict the label. Furthermore, figure also shows how the model output (predicted probability) is calibrated with the performance of the model. Though these findings show the utility of our dataset and model in efficiently automating the detection of viable samples (a labor-intensive task), reducing the required detection time remains an open challenge.

\begin{figure}[t]
\includegraphics[width=0.8\textwidth]{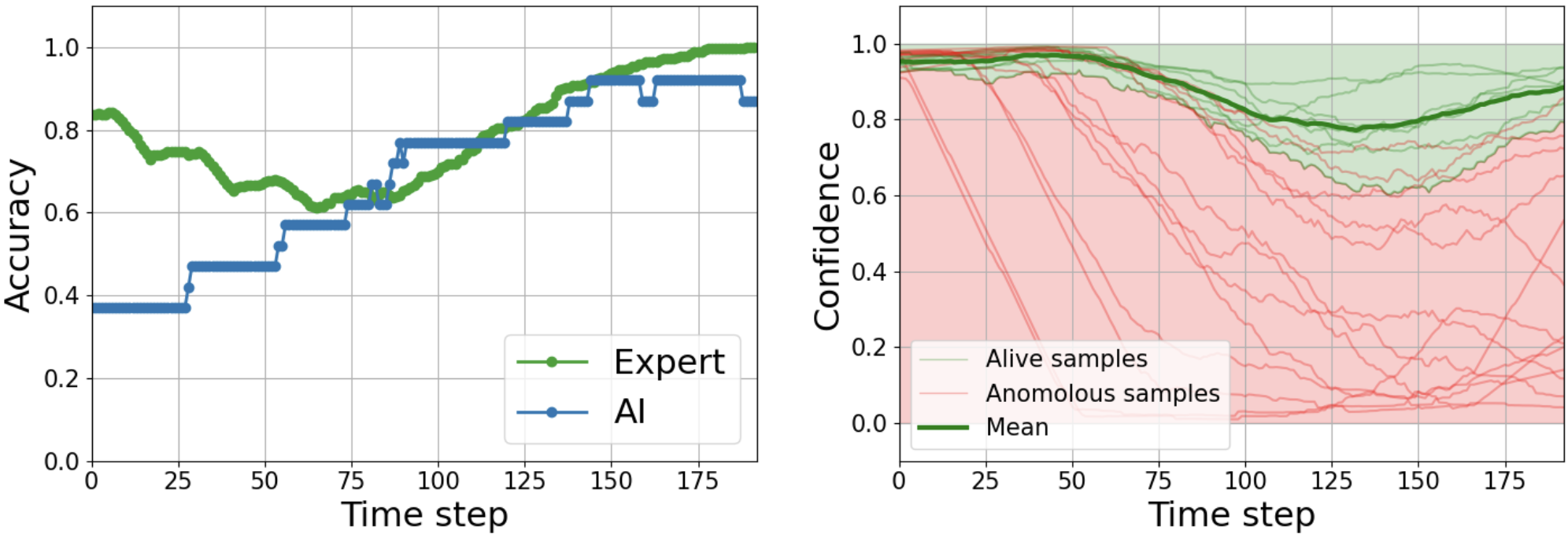}
\centering
\caption{The figure (left) shows the comparison of the accuracy of human prediction to that of the model prediction at all time instances. The second figure shows the confidence in prediction varying over time for samples of the two classes.} 
\Description{Left plot contrasts human versus model accuracy across all time points; right plot shows confidence trajectories diverging for ‘alive’ and ‘anomalous’ sequences as observation time increases.}
\label{result2}
\end{figure}

For the second task-toxicity detection-we have a total of $55,296$ images, constituting $288$ sequences. Among these, $143$ sequences exhibit normal development while $112$ sequences display developmental anomalies due to toxic exposure (the remaining samples are not fertile). The train test validation splits follow the same ratio as the previous task. The model’s prediction is obtained by taking the \textit{argmax} over the output logits for the $2$ classes and the corresponding logit is the confidence in prediction. 

Experts observe the samples sequentially, assigning a label of ‘alive’ to early images. Since the true sequence label is provided at the end of each sequence, we can evaluate how experts can accurately predict early labels. The results show that the accuracy of model prediction also improves with time. Figure~\ref{result2} shows the comparison of the accuracy of human prediction to that of the model prediction at all time instances. There is a clear scope for improvement for the model to match human performance of early accurate prediction (best performance is 92\%). The figure also shows the confidence in prediction which can be dynamically thresholded to make a decision at each time instance. The threshold for each time step can be optimized with the validation data. These findings demonstrate that while both human experts and our model improve in accuracy over time, there remains significant potential to refine the automated approach for earlier and more precise toxicity detection.

\section {Conclusion}
\label{sec:conclusion}
In conclusion, our work presents a comprehensive framework for automating developmental toxicity screening in zebrafish embryos, bridging a critical gap in current drug discovery pipelines. By introducing a large-scale, finely annotated dataset and a transformer-based model, we demonstrate the feasibility of robust, automated early detection of developmental toxicity with high accuracy. We believe, the annotated dataset will be a valuable benchmark for biomedical research. Better models can be developed on this dataset for early-stage detection which is currently at around $3.4$ hours after development. 

\begin{credits}
\subsubsection{\ackname}%
This work was partially funded by Image-Tox (ZT-I-PF-4-037), supported by the impulse and networking fund of the Helmholtz Association. The views and opinions expressed are those of the authors only and do not necessarily reflect those of the funding agencies, which can neither be held responsible for them. It was also partially funded by ELSA – European Lighthouse on Secure and Safe AI (grant agreement No. 101070617). The project on which this report is based was funded by the German Federal Ministry of Education and Research (funding code 16KIS2012).

\subsubsection{\discintname}%
The authors have no competing interests to declare that are relevant to the content of this article.
\end{credits}

\bibliographystyle{splncs04}
\bibliography{mybibliography}

\begin{thebibliography}{10}
\providecommand{\url}[1]{\texttt{#1}}
\providecommand{\urlprefix}{URL }
\providecommand{\doi}[1]{https://doi.org/#1}

\bibitem{alshut2010methods}
Alshut, R., Legradi, J., Liebel, U., Yang, L., van Wezel, J., Str{\"a}hle, U., Mikut, R., Reischl, M.: Methods for automated high-throughput toxicity testing using zebrafish embryos. In: KI 2010: Advances in Artificial Intelligence: 33rd Annual German Conference on AI, Karlsruhe, Germany, September 21-24, 2010. Proceedings 33. pp. 219--226. Springer (2010)

\bibitem{cachat2011three}
Cachat, J., Stewart, A., Utterback, E., Hart, P., Gaikwad, S., Wong, K., Kyzar, E., Wu, N., Kalueff, A.V.: Three-dimensional neurophenotyping of adult zebrafish behavior. PloS one  \textbf{6}(3),  e17597 (2011)

\bibitem{cario2011automated}
Cario, C.L., Farrell, T.C., Milanese, C., Burton, E.A.: Automated measurement of zebrafish larval movement. The Journal of physiology  \textbf{589}(15),  3703--3708 (2011)

\bibitem{dong2015deep}
Dong, B., Shao, L., Da~Costa, M., Bandmann, O., Frangi, A.F.: Deep learning for automatic cell detection in wide-field microscopy zebrafish images. In: 2015 IEEE 12th international symposium on biomedical imaging (ISBI). pp. 772--776. IEEE (2015)

\bibitem{dosovitskiy2020image}
Dosovitskiy, A.: An image is worth 16x16 words: Transformers for image recognition at scale. arXiv preprint arXiv:2010.11929  (2020)

\bibitem{horzmann2018making}
Horzmann, K.A., Freeman, J.L.: Making waves: New developments in toxicology with the zebrafish. Toxicological Sciences  \textbf{163}(1),  5--12 (2018)

\bibitem{iakovidis2018detecting}
Iakovidis, D.K., Georgakopoulos, S.V., Vasilakakis, M., Koulaouzidis, A., Plagianakos, V.P.: Detecting and locating gastrointestinal anomalies using deep learning and iterative cluster unification. IEEE transactions on medical imaging  (2018)

\bibitem{javanmardi2023unsupervised}
Javanmardi, S., Tang, X., Jahanbanifard, M., Verbeek, F.J.: Unsupervised segmentation of high-throughput zebrafish images using deep neural networks and transformers. In: International Conference on Data Science and Artificial Intelligence. pp. 213--227. Springer (2023)

\bibitem{karpathy2014large}
Karpathy, A., Toderici, G., Shetty, S., Leung, T., Sukthankar, R., Fei-Fei, L.: Large-scale video classification with convolutional neural networks. In: Proceedings of the IEEE Conference on Computer Vision and Pattern Recognition (CVPR) (2014)

\bibitem{kokel2010rapid}
Kokel, D., Bryan, J., Laggner, C., White, R., Cheung, C.Y.J., Mateus, R., Healey, D., Kim, S., Werdich, A.A., Haggarty, S.J., et~al.: Rapid behavior-based identification of neuroactive small molecules in the zebrafish. Nature chemical biology  \textbf{6}(3),  231--237 (2010)

\bibitem{lieschke2007animal}
Lieschke, G.J., Currie, P.D.: Animal models of human disease: zebrafish swim into view. Nature Reviews Genetics  \textbf{8}(5),  353--367 (2007)

\bibitem{macrae2015zebrafish}
MacRae, C.A., Peterson, R.T.: Zebrafish as tools for drug discovery. Nature reviews Drug discovery  \textbf{14}(10),  721--731 (2015)

\bibitem{mikut2013automated}
Mikut, R., Dickmeis, T., Driever, W., Geurts, P., Hamprecht, F.A., Kausler, B.X., Ledesma-Carbayo, M.J., Mar{\'e}e, R., Mikula, K., Pantazis, P., et~al.: Automated processing of zebrafish imaging data: a survey. Zebrafish  \textbf{10}(3),  401--421 (2013)

\bibitem{mirzaei2022fake}
Mirzaei, H., Salehi, M., Shahabi, S., Gavves, E., Snoek, C.G., Sabokrou, M., Rohban, M.H.: Fake it until you make it: Towards accurate near-distribution novelty detection. In: The eleventh international conference on learning representations (2022)

\bibitem{5872696}
Ohn, J., Liebling, M.: In vivo, high-throughput imaging for functional characterization of the embryonic zebrafish heart. In: 2011 IEEE International Symposium on Biomedical Imaging: From Nano to Macro. pp. 1549--1552 (2011). \doi{10.1109/ISBI.2011.5872696}

\bibitem{padilla2012zebrafish}
Padilla, S., Corum, D., Padnos, B., Hunter, D., Beam, A., Houck, K., Sipes, N., Kleinstreuer, N., Knudsen, T., Dix, D., et~al.: Zebrafish developmental screening of the toxcast™ phase i chemical library. Reproductive toxicology  \textbf{33}(2),  174--187 (2012)

\bibitem{rennekamp201515}
Rennekamp, A.J., Peterson, R.T.: 15 years of zebrafish chemical screening. Current opinion in chemical biology  \textbf{24},  58--70 (2015)

\bibitem{rihel2010zebrafish}
Rihel, J., Prober, D.A., Arvanites, A., Lam, K., Zimmerman, S., Jang, S., Haggarty, S.J., Kokel, D., Rubin, L.L., Peterson, R.T., et~al.: Zebrafish behavioral profiling links drugs to biological targets and rest/wake regulation. Science  \textbf{327}(5963),  348--351 (2010)

\bibitem{schunck2024integrating}
Schunck, F., Kodritsch, B., Krauss, M., Busch, W., Focks, A.: Integrating time-resolved nrf2 gene-expression data into a full guts model as a proxy for toxicodynamic damage in zebrafish embryo. Environmental Science \& Technology  \textbf{58}(50),  21942--21953 (2024)

\bibitem{spomer2012high}
Spomer, W., Pfriem, A., Alshut, R., Just, S., Pylatiuk, C.: High-throughput screening of zebrafish embryos using automated heart detection and imaging. Journal of laboratory automation  \textbf{17}(6),  435--442 (2012)

\bibitem{truong2014multidimensional}
Truong, L., Reif, D.M., St~Mary, L., Geier, M.C., Truong, H.D., Tanguay, R.L.: Multidimensional in vivo hazard assessment using zebrafish. Toxicological sciences  \textbf{137}(1),  212--233 (2014)

\bibitem{tyagi2018fine}
Tyagi, G., Patel, N., Sethi, I.: A fine-tuned convolution neural network based approach for phenotype classification of zebrafish embryo. Procedia Computer Science  \textbf{126},  1138--1144 (2018)

\bibitem{vaswani2017attention}
Vaswani, A., Shazeer, N., Parmar, N., Uszkoreit, J., Jones, L., Gomez, A.N., Kaiser, L., Polosukhin, I.: Attention is all you need.(nips), 2017. arXiv preprint arXiv:1706.03762  \textbf{10},  S0140525X16001837 (2017)

\bibitem{xu2024technical}
Xu, L., Wang, S.: Technical report: Masked skeleton sequence modeling for learning larval zebrafish behavior latent embeddings. arXiv preprint arXiv:2403.15693  (2024)

\bibitem{xu2017zebrafish}
Xu, Z., Cheng, X.E.: Zebrafish tracking using convolutional neural networks. Scientific reports  \textbf{7}(1),  42815 (2017)

\bibitem{zhang2017new}
Zhang, G., Truong, L., Tanguay, R.L., Reif, D.M.: A new statistical approach to characterize chemical-elicited behavioral effects in high-throughput studies using zebrafish. PloS one  \textbf{12}(1),  e0169408 (2017)

\bibitem{zhao2021good}
Zhao, Y., Wu, W., He, Y., Li, Y., Tan, X., Chen, S.: Good practices and a strong baseline for traffic anomaly detection. In: Proceedings of the IEEE/CVF Conference on computer vision and pattern recognition (2021)

\bibitem{zon2005vivo}
Zon, L.I., Peterson, R.T.: In vivo drug discovery in the zebrafish. Nature reviews Drug discovery  \textbf{4}(1),  35--44 (2005)

\end{thebibliography}
\end{document}